\documentclass[11pt]{article}
\usepackage[margin=1in]{geometry}
\usepackage{graphicx}
\usepackage{booktabs}
\usepackage{amsmath}
\usepackage[hidelinks]{hyperref}
\usepackage{xcolor}
\usepackage{caption}
\title{Do Ray-Traced Cities Make a Better Channel Prior for an Unseen Site?\\[4pt]
\large A pre-registered study in simulation, and a test on measured channels that reverses it}
\author{Krish Patel\\
\normalsize University of Waterloo\\
\normalsize \href{mailto:k77patel@uwaterloo.ca}{\texttt{k77patel@uwaterloo.ca}}}
\date{November 2026}

\begin{document}
\maketitle

\begin{abstract}
Channel estimators at a newly deployed base station usually start from 3GPP
stochastic channel models, which describe no particular street. Ray tracers
can simulate real city layouts instead. We ask whether a channel prior pooled
from many ray-traced cities serves an \emph{unseen} site better than the 3GPP
CDL models, and whether learned estimators are what make the difference. With
Sionna RT on 14 city scenes, LMMSE estimation from covariance priors,
leave-one-city-out evaluation, and outcomes and pass/fail bars written down
before each main experiment, we find in simulation that the pooled prior sits
about 0.6\,dB from a site-trained estimator with no site measurements---about
what CDL reaches after 50 measurements---and protects against a badly matched
source city. Learned models added little: at most 0.65\,dB in estimation, and
in channel prediction a simple per-user classical method beat CNN predictors
by 1.8--12\,dB from 2\,ms ahead. On measured channels from the DICHASUS outdoor
dataset (1.272\,GHz, three 2$\times$8 arrays on one building) the result
reversed: the pooled ray-traced prior was 3.2--4.1\,dB \emph{worse} than CDL.
The simulated channels were too clean. Enabling diffuse scattering nearly
closed the gap (to 0.3--0.5\,dB) only while simulated users stood farther away
than the real ones; with users at the real distances, even a scattering level
tuned from 50 site measurements left the prior about 2\,dB behind CDL. The map-built scenes miss
multipath the real street has. A pre-registered follow-on on two more measured
environments found the generic CDL prior 2.1--4.7\,dB from a site-trained one
on the full grid:
no further in a factory hall than on the street, furthest in a small lab room
whose delay spread sits at the low end of the prior's range. Finally, a hallway
model calibrated to measurements by differentiable ray tracing reproduced delay
spread, but as a prior in an unmeasured part of the hallway it was 4.5\,dB
worse than CDL on the full array. All measured results come from one sounder.
\end{abstract}

\section{Introduction}

A base station estimates the radio channel to each user from a few pilot
symbols. Linear MMSE (LMMSE) estimation does this well when it knows the
channel's second-order statistics (its covariance across antennas, subcarriers
and time); learned estimators can in principle exploit more. At a new site,
neither the covariance nor training data is available, so estimators start from
a generic prior---typically the 3GPP CDL/TDL models of TR~38.901~\cite{tr38901}---and
adapt as site measurements arrive.

Differentiable ray tracers such as Sionna RT~\cite{sionnart} can simulate
propagation in real city geometry. Recent work uses ray-traced data as a
site-specific prior for a known site
\cite{delmoro2026,alikhani2025,abouamer2024}, pre-trains across source cities for
CSI feedback~\cite{sifo2026}, and compares stochastic with ray-traced data for
machine learning~\cite{morais2025}. We ask a narrower question: for a site
\emph{with no geometry model and no measurements}, is a prior pooled from
\emph{other} ray-traced cities better than the 3GPP models? And is a learned
estimator what makes the difference?

We answer in three steps, plus two follow-ons: (i) leave-one-city-out experiments in simulation,
(ii) the same question for short-term channel prediction, where learning has
more room, and (iii) a test of the main simulation finding on measured
channels; the follow-on (iv) asks how far the 3GPP prior falls short in three
measured environments, and a final trial (v) whether a ray-traced model
calibrated to measurements does better. Each main experiment's expected outcome and pass/fail bar was
committed to the lab notebook before it ran; follow-up checks are labelled as
post-hoc or exploratory. The code, notebook, logs and result tables are
available from the author on request.

Our contributions are: (1) a pre-registered leave-one-city-out comparison of
ray-traced and 3GPP priors across 14 cities, showing pooling as insurance in
simulation; (2) evidence that, on this simulator, classical estimators and
predictors match or beat learned ones; (3) a test of the simulation finding on
measured channels with one processing chain for every source, which reverses
it; (4) a diagnosis consistent with the data: the gap comes from
multipath the map-built scenes lack, which neither more cities, scattering, nor a tuned scattering level
recovers; and (5) a measurement of how far the generic 3GPP prior falls short
in three environments, which found it no larger on the full grid in a factory hall than on an
open street.

\section{Setup}

\paragraph{Grid and estimation.} A 5G NR-like OFDM slot: 3.5\,GHz carrier,
30\,kHz subcarrier spacing, 128 subcarriers $\times$ 14 symbols, DMRS-like
pilots on every other subcarrier of symbols 2 and 11, and an 8-antenna
half-wavelength uniform linear array at the base station. Channels are
normalised to unit power per sample. The estimator is LMMSE from a covariance
prior $(R_{hp}, R_{pp})$ estimated from simulated channels (12k per source in
simulation; 6000 per city on measured data, 2000 in the exploratory runs). With $N$ site
measurements, the prior is shrunk toward their sample covariance with a weight
chosen by 5-fold cross-validation. A residual CNN (zero-initialised last layer,
so it starts at its LMMSE front end) tests whether anything beyond second-order
statistics helps. Error is NMSE at 10\,dB SNR unless stated, and the
``ceiling'' is LMMSE from 12k channels of the target site itself.

\paragraph{Channel sources.} All channels are generated with
Sionna~\cite{sionna}. \emph{CDL}: 3GPP CDL-A to E with delay spread
30--300\,ns and a random base-station heading per sample (a fixed heading made
CDL artificially poor in an early run, which we corrected and report).
\emph{Ray-traced (RT)}: Sionna RT~2.2 with up to 3 specular bounces and
refraction, in Sionna's four built-in scenes (Munich, Paris-\'Etoile, Florence,
San Francisco) and ten New York State areas built from OpenStreetMap with
geo2sigmap~\cite{geo2sigmap}, reading buildings from an offline Geofabrik
extract. Base stations are placed 25--40\,m up near the scene centre and users
within 250\,m.

\paragraph{Protocol.} Each city in turn is the unseen target; priors come from
the other cities (pooled with equal weight) or from CDL. Test users are at
base-station sites the priors never saw. The realism result was re-checked on
fresh base-station sites and freshly drawn pre-training sets.

\section{Results in simulation}

\paragraph{Ray-traced vs CDL, four cities.} With no site data, a prior pooled
from the three other built-in cities beat CDL in Munich (+0.3--0.5\,dB),
Florence (+1.1--1.2) and San Francisco (+1.3--1.5) and lost in \'Etoile
(--1.2 to --1.8) across the original and two fresh draws of sites and
pre-training data. The advantage faded with site data: by $N{=}1000$ all priors
were within $\sim$0.2\,dB except \'Etoile.

\begin{figure}[t]
\centering
\includegraphics[width=0.68\linewidth]{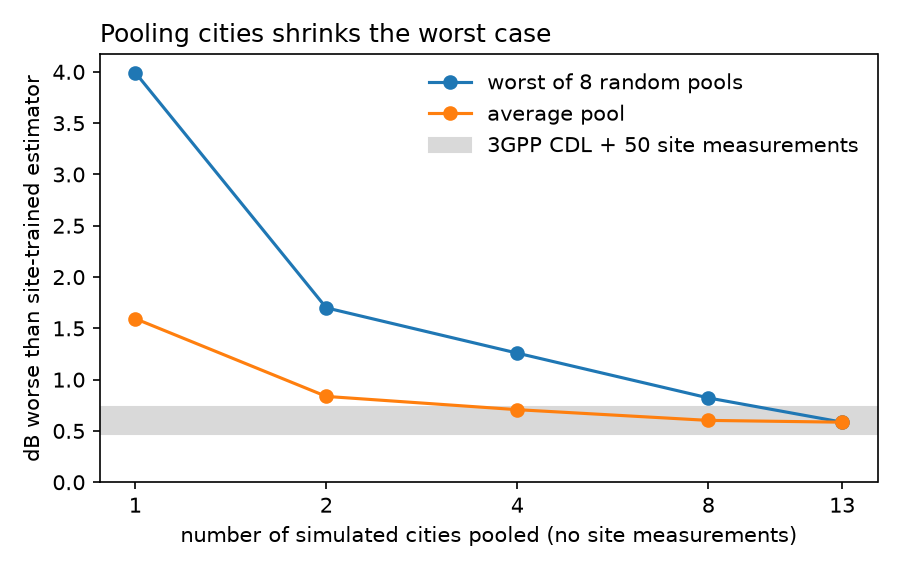}
\caption{Simulation: gap to a site-trained estimator vs.\ the number of pooled
ray-traced cities, no site measurements, mean over 14 targets. The grey band is
3GPP CDL after blending in 50 site measurements.}
\label{fig:breadth}
\end{figure}

\paragraph{Fourteen cities: pooling as insurance.} Pooling $k$ random other
cities (8 draws each), the \emph{worst} draw's gap to the ceiling shrank from
3.99\,dB at $k{=}1$ to 0.82\,dB at $k{=}8$, while the average went from 1.59 to
0.60\,dB (Fig.~\ref{fig:breadth}): pooling protects against a badly matched
source city (the worst single-city cases, 6--7\,dB, all used a nearly empty
open-area scene as the source for a dense downtown). The all-cities pool beat
CDL in 11 of 14 targets by 0.29\,dB on average. With no site data it sat
0.59\,dB from the ceiling, about where CDL lands after 50 site measurements
(0.47--0.73\,dB across four draws of those 50; post-hoc). Once 50
measurements are blended in, the two are comparable (pooled ahead by
$\sim$0.1\,dB on built-up sites; post-hoc). Choosing a single best source city from 50 measurements worked with 3
candidates (mean regret 0.05\,dB) but overfit with 15 (0.48\,dB, against a
pre-registered bar of 0.3).

\paragraph{Learning in estimation.} The CNN refiner added $\sim$0.2--0.45\,dB
on the CDL prior and often nothing on ray-traced priors, at most 0.65\,dB in any
cell (one seed). The covariance did almost all of the work.

\section{Learning where it should have room: channel prediction}

Predicting the channel 0.5--10\,ms ahead from $\sim$1.8\,ms of noisy pilot history
(8 observations over 4 slots) is a task where per-user adaptation matters: a
covariance pooled over users cannot know each user's Doppler. Sionna RT
evolves a moving user by rotating the phases of a fixed set of paths; we
checked that this matches re-tracing at the moved position up to $\sim$10\,ms
for nearly all users at 3.5\,GHz (with the solver's deterministic mode), so we
restricted horizons to 10\,ms.

\begin{figure}[t]
\centering
\includegraphics[width=0.68\linewidth]{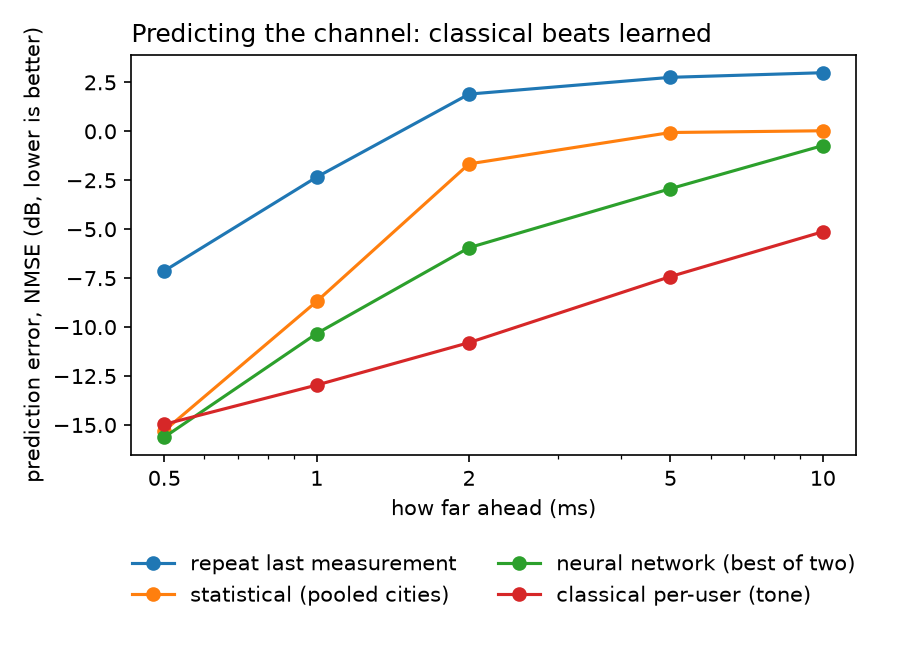}
\caption{Simulation: prediction error vs.\ horizon, mean over four cities,
10\,dB SNR, no site data. ``Neural network'' is the better of the two CNN arms
in each cell.}
\label{fig:pred}
\end{figure}

On 3GPP-compliant simulated channels, learned predictors have been reported to
beat a Prony-based angle--delay baseline~\cite{wifo}; neural predictors
have also been compared on the 3GPP TDL model~\cite{stenhammar2023}. We compared repeating the last observation, space-time Wiener prediction from
covariance priors, a classical per-user method (``tone'': in the angle--delay
domain, fit one Doppler tone per bin to the history and extrapolate), and CNN
predictors in the antenna and angle--delay domains. The pre-registered test for
``learning has room'' failed: tone beat both CNN arms at every horizon from
2\,ms in every city, by 1.8--4.7\,dB in Munich, \'Etoile and Florence and by
$\sim$9--12\,dB in San Francisco (Fig.~\ref{fig:pred}). The CNNs did beat
covariance-based prediction by 3--6\,dB at 2\,ms; learning was competitive only
one slot (0.5\,ms) ahead.

A post-hoc follow-up trained a CNN to \emph{correct} tone's prediction, with a
loss weighted toward the predictable horizons. It added 0.3--0.6\,dB at
2--10\,ms in three of four cities (cells 0.26--0.65; two seeds agree within
$\sim$0.2\,dB) and cost 0.8--1.2\,dB at 5--10\,ms in San Francisco, where tone is
already near-exact. Re-weighting the original CNNs' loss alone did not help.
The caveat that matters most: fixed-path evolution is tone's own model, so on
real channels tone would degrade by an amount this simulator cannot measure.

\section{Test on measured channels}
\label{sec:measured}

\paragraph{Data.} DICHASUS~\cite{dichasus} outdoor set \texttt{dxxx}
\cite{dichasusdata}: three 2$\times$8 arrays ($\lambda/2$ at 1.272\,GHz) on one
building facade about 10\,m up and tilted $\sim$20$^\circ$ down, a hand-pulled
transmitter on the street, 50\,MHz over 1024 subcarriers, tachymeter
positions. Calibration files (also used for the ceiling), a site file (50
random measurements, 4 draws) and test files are disjoint recordings.

\paragraph{Processing.} The receivers carry constant per-antenna phase and
timing offsets, as the dataset authors describe; offset files are not shipped
for this set. We fit, per antenna, a phase plus a delay relative to a reference
antenna to the line-of-sight phase predicted from the tachymeter positions;
the fit also identified the array layout (8-element axis horizontal: coherence
0.73 on array A versus at most 0.32 for the alternative layouts in an early fit;
0.80 on the full calibration set).
Every source---measured, CDL and ray-traced---is generated on the full
50\,MHz band and passes through one chain: align the first arrival to delay
zero (measured datapoints are first shifted so their strongest tap sits at
zero, removing the recording's timing offset; added during the follow-on, it
leaves the street results unchanged), keep a delay window of $-$200\,ns to 2\,$\mu$s, take the 128 subcarriers
around the carrier (6.25\,MHz), normalise. The grid is 16 antennas $\times$ 128
subcarriers in one snapshot, pilots on every other subcarrier. All 14 cities
were re-simulated at 1.272\,GHz with the 2$\times$8 array, a directional
38.901 element pattern, a 20$^\circ$ downtilt and a facade-like placement
(base station 10\,m above ground, users 10--65\,m in front, outdoor positions
only, heights following terrain).

\begin{figure}[t]
\centering
\includegraphics[width=0.68\linewidth]{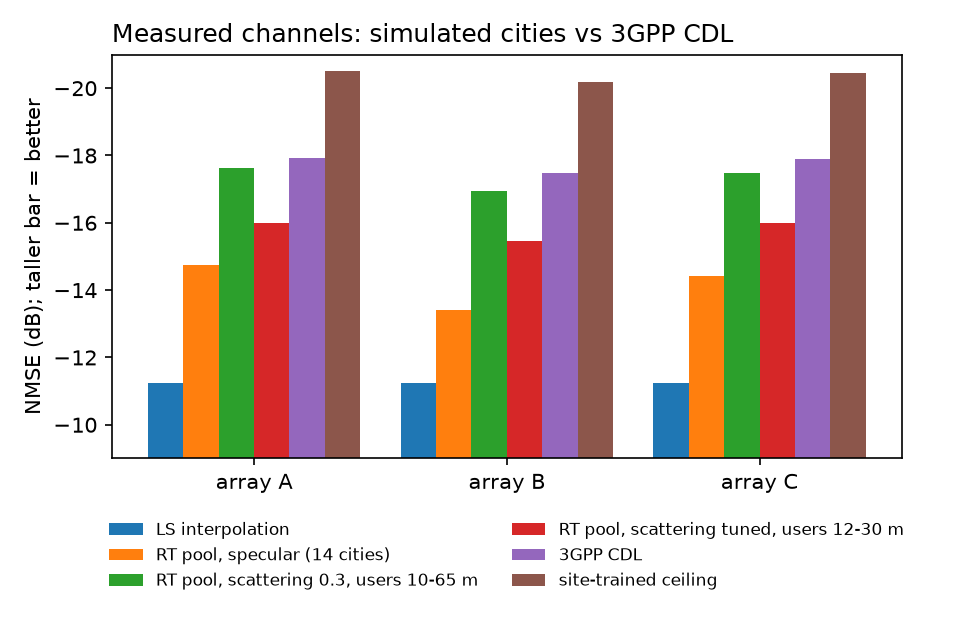}
\caption{Measured channels, the same numbers as Table~\ref{tab:measured};
taller bars are better.}
\label{fig:measured}
\end{figure}

\paragraph{Results.} The pre-registered claim that the pooled ray-traced prior
beats CDL with no site data was falsified on all three arrays
(Table~\ref{tab:measured}, Fig.~\ref{fig:measured}): it was 3.2--4.1\,dB worse
on the full array $\times$ frequency grid and 0.5--1.3\,dB worse per antenna.
On the full grid no single city beat CDL either. With 50 site measurements CDL
still led by 1.2--1.7\,dB on the full grid; per antenna both reached the
site-trained level. Pooling still shrank the worst single-city case in 5 of 6
checks, but around a level below CDL. The pool sat 5.8--6.8\,dB from the
ceiling, against $\sim$0.6\,dB in simulation.

\begin{table}[t]
\centering
\small
\begin{tabular}{lrrrrrr}
\toprule
array & LS interp. & RT specular & RT scatt., far users & RT tuned, real distances & CDL & ceiling \\
\midrule
A & $-$11.2 & $-$14.7 & $-$17.6 & $-$16.0 & $-$17.9 & $-$20.5 \\
B & $-$11.2 & $-$13.4 & $-$17.0 & $-$15.5 & $-$17.5 & $-$20.2 \\
C & $-$11.2 & $-$14.4 & $-$17.5 & $-$16.0 & $-$17.9 & $-$20.4 \\
\bottomrule
\end{tabular}
\caption{Measured DICHASUS channels: NMSE (dB) at 10\,dB SNR on the full
16-antenna $\times$ 128-subcarrier grid. Specular: 14-city pool, users
10--65\,m. Scattering, far users: 7 cities, scattering 0.3, users 10--65\,m.
Tuned, real distances: 7 cities, users 12--30\,m, scattering level picked by
50 site measurements (the only prior column that uses site data). The last two are
exploratory.}
\label{tab:measured}
\end{table}

\paragraph{Why.} The simulated channels were cleaner in delay than the
measured street: 89--100\% of energy in the first 160\,ns tap across the 14
cities, against 79--84\% measured (CDL 78\%). Spatial concentration was not
clearly different. The ray tracer ran with the original study's settings:
specular reflection and refraction only, no diffuse scattering, no diffraction,
every material's scattering coefficient zero. In an exploratory rerun, added
after a dry run and labelled as such, we enabled diffuse scattering
(coefficient 0.3, Lambertian) and Sionna's wedge diffraction for 7 cities.
The pooled prior came within 0.3--0.5\,dB of CDL on the full grid and level
with it per antenna (0.3\,dB ahead on arrays B and C). A control with the same cities and the same number of
base-station sites but no scattering attributes $\sim$2.8\,dB of the
$\sim$3\,dB improvement on array A to scattering and diffraction and
$\sim$0.2\,dB to the extra sites.

\paragraph{Matching the users' distance made it worse.} The simulated users
sat farther from the array than the real ones (median 46\,m vs 18--20\,m). In a
second exploratory probe, with scattering on and users placed 12--30\,m away,
the ray-traced prior fell to 2.9\,dB behind CDL on array A (2.4--2.6\,dB worse
than the farther placement on the full grid of every array, 1.5--2.2\,dB per
antenna). Closer simulated users
are more line-of-sight dominated (first-tap share 89--97\%), while the real
users at those distances show 79--84\%: the measured street has multipath
the map-built scenes do not produce, plausibly from objects they lack
(vehicles, trees, street furniture, facade detail). The near-match of the
scattering variant therefore came partly from placing users farther away than
the real ones.

\paragraph{Tuning the scattering level from 50 site measurements.} With users at
the measured distances, we simulated the 7 cities at diffuse-scattering
coefficients 0, 0.15, 0.3, 0.5 and 0.7 and let each draw of 50 site
measurements pick the level whose pooled prior scored best on them. On the
full grid the measurements always picked 0.7, which was also best on the test
data, but the
tuned prior stayed 1.9--2.0\,dB behind CDL on the full grid of every array
(0.9--1.5\,dB behind CDL\,+\,50 once the 50 measurements were blended in), and
per antenna the scattering level made no difference (1.2--2.3\,dB behind
CDL). Diffuse scattering off the modelled buildings does not supply the delay
spread the street has at short range; the related work points to street
clutter (vehicles, poles, trees) and to 3GPP-style random clusters on top of
the ray-traced paths~\cite{tr38901}.

\section{Where does CDL fall short? Three measured environments}
\label{sec:cdlgap}

A site-specific prior, ray-traced or otherwise, only earns its cost where the
3GPP models are a poor prior. As a follow-on, pre-registered before the
lab and factory results (the street gap was already known), we measured that gap in three environments recorded with the same
sounder at 1.272\,GHz: the street above, an indoor lab room with two 2$\times$8
arrays (\texttt{0d5x}~\cite{dichasuslab}) and a factory hall with four
2$\times$4 arrays placed around it, with line-of-sight and blocked positions
(\texttt{cf0x}~\cite{dichasusfactory}).

\paragraph{Design.} The estimator, grid, calibration and processing chain are
those of Section~\ref{sec:measured}, with no downtilt indoors. Indoors, the
site-trained ceiling is fitted on 80\% of one recording, the 50 site
measurements (4 draws) come from the remaining contiguous 20\%, and testing
uses a different recording. All arrays passed the pre-set calibration bar
(coherence $\geq$0.5): 0.71 and 0.81 in the lab, 0.85--0.91 in the factory.
The lab's positions are two-dimensional; the transmitter height was set from
calibration coherence alone, before any estimation result. CDL priors were
re-simulated for each array shape. Besides the generic CDL mixture (delay
spreads drawn from 30--300\,ns), we scored each of 15 fixed profiles (CDL-A to
E at 30, 100 and 300\,ns) and the profile that the 50 site measurements pick.
Three claims were registered: (G1) the generic-CDL gap to the ceiling is
larger in the factory than on the street (full grid, mean over arrays); (G2)
if the gap is below 1\,dB everywhere, CDL is good enough and we stop; (G3) a
profile picked from 50 site measurements closes at least half of the gap
wherever it is at least 1\,dB.

\begin{table}[t]
\centering
\footnotesize
\setlength{\tabcolsep}{4pt}
\begin{tabular}{lccccc}
\toprule
 & & \multicolumn{4}{c}{dB above the site-trained ceiling: full grid / per antenna} \\
\cmidrule(l){3-6}
environment & first tap & generic CDL & CDL\,+\,50 & picked & oracle \\
\midrule
street (3 arrays, 2$\times$8) & 79--84\% & 2.5--2.7 / 0.8--0.9 & 1.4--1.7 / 0.0--0.1 & 4.8--5.6 / 0.3--0.6 & 4.8--5.6 / 0.3--0.6 \\
lab room (2, 2$\times$8) & 95--96\% & 4.7 / 4.3 & 1.5--2.5 / 0.0 & 2.3--2.5 / 0.4 & 2.3--2.5 / 0.4 \\
factory (4, 2$\times$4) & 81--94\% & 2.1--2.8 / 0.7--1.4 & 1.0--2.5 / 0.2--1.1 & 3.5--6.8 / 0.5--1.9 & 3.3--4.5 / 0.3--0.5 \\
\bottomrule
\end{tabular}
\caption{The CDL gap in three measured environments at 10\,dB SNR: ranges over
arrays, means over the four draws of 50 site measurements. First tap: share of
energy in the first 160\,ns delay tap of the 6.25\,MHz grid (CDL's own:
$\sim$78\%). Picked profile: the single fixed CDL profile the 50 site
measurements pick; oracle: the best one on the test recording.}
\label{tab:cdlgap}
\end{table}

\paragraph{Results.} These are the results after a processing fix made
during review (described at the end of this section); the first, faulty
analysis had supported G1. G1 is falsified: the factory gap (2.1--2.8\,dB,
mean 2.4) is no larger than the street's (2.5--2.7, mean 2.6), also at 0\,dB
SNR (Table~\ref{tab:cdlgap}); per antenna, a secondary reading, it is
slightly larger (1.1 vs 0.8\,dB). The stop rule G2 was not triggered: the
full-grid gap is at least 2\,dB everywhere. The largest gap is in the lab room
(4.7\,dB full grid, 4.3 per antenna). G3 named no basis. On the full grid, the
basis of G1 and G2, it is met nowhere: a single picked profile was
\emph{worse} than the generic mixture on the street and in the factory, and
closed 49\% of the gap in the lab (53\% at 0\,dB SNR, just over the bar). Per
antenna it is met in the lab (4.3 to 0.4\,dB) but not in the factory, where
picking helped on arrays B and D and hurt on A and C (mean 1.09 to 1.06\,dB).
Blending the 50 site measurements into the generic prior helped most: at
10\,dB SNR, within 0.2\,dB of the ceiling per antenna on the street and in the
lab, at most 1.1\,dB in the factory, and 1.0--2.5\,dB on the full grid. It was
not uniformly helpful. Indoors the 50 come from one stretch of the recording;
on factory arrays A and C that stretch sat about 4\,m from the array against
11--12\,m for the test recording, and there, per antenna, the 50 mostly picked
a line-of-sight profile (CDL-E) and blending helped little or hurt (0.84 to
0.69\,dB; 0.71 to 1.10). On arrays B and D the mismatch ran the other way
(11--13 vs 5\,m) and blending worked, so distance alone does not explain the
difference. At 0\,dB SNR blending hurt per antenna in the factory on average
(0.43 to 0.59\,dB). The pre-registration also listed RMS delay spread and
spatial eigenvalue shares, to be reported without a bar; we dropped them,
because on this grid the RMS statistic is dominated by leakage and noise and
does not track the true delay spread, and the spatial share is not comparable
between 8- and 16-antenna arrays. The factory test recording was not split
into line-of-sight and blocked positions.

\paragraph{Reading.} The pre-registered expectation, that CDL falls further
short in a dense indoor space than on an open street, is not supported by
these recordings: the factory hall is about as close as the street. The
outlier is the small lab room, whose energy nearly all arrives within one
160\,ns tap, at the low end of the generic mixture's 30--300\,ns delay-spread
range. That is partly a choice of the mixture rather than a failure of the
3GPP models: TR~38.901 scales CDL to a scenario's delay spread, with examples
from 10 to 1000\,ns~\cite{tr38901}, and a 30\,ns profile fixes the lab per
antenna (on the full grid it closes about half the gap). It is not the whole
story: factory arrays B and D also put 93--94\% of their energy in the first
tap but show much smaller gaps, for reasons we have not established. None of
the three environments stands out as one where a site-specific prior would
clearly pay off. Three recordings with different arrays cannot rule out such
sites elsewhere, and this study did not test a ray-traced prior indoors.

\paragraph{A correction.} A review of this follow-on found that the factory
recordings carry a timing offset that drifts across the whole delay range
between datapoints, far beyond the chain's first-arrival search. A first
analysis therefore kept mostly noise for the factory and reported a
10--12\,dB gap, which supported G1; a post-hoc check with longer-delay CDL
profiles built on it was withdrawn. Measured channels are now first shifted so
that each datapoint's strongest tap sits at delay zero
(Section~\ref{sec:measured}). A synthetic check fails without this shift and
passes with it; no street or lab datapoint changed by more than $10^{-6}$; and
in no factory datapoint does a tap within 10\,dB of the strongest arrive 400\,ns to 2\,$\mu$s
before it, where the chain's search would miss it. The numbers
above are the rerun.

\section{A calibrated ray-traced hallway}\label{sec:hallway}

The street scenes were built from maps and never fitted to the site. A
pre-registered final trial asked whether a scene fitted to measurements does
better. DICHASUS \texttt{dcxx}~\cite{dichasushallway} (recording
\texttt{dc01}, 3.438\,GHz, 50\,MHz) observes a robot-carried dipole in an
institute hallway from two 4$\times$8 arrays at opposite ends; its simplified
3D model (floor, walls, ceiling, one material each) is the one calibrated by
differentiable ray tracing in~\cite{hoydis2023learning}. We first reproduced
that calibration (permittivity and conductivity of the three objects, fitted to
per-antenna power and RMS delay spread): held-out delay-spread error fell from
0.42 to 0.28 (paper: 0.54 to 0.24), and power error fell in the corridor
(4.2 to 2.4\,dB) but rose in the lobby (11.0 to 16.0\,dB), where the model is
too simple. The calibration was then refitted on one part of the corridor
($y<0$) and the model used as a prior in another ($y>10$\,m; the band between,
which holds the lobby, was left out): channels simulated at 12{,}000 positions over that
part, with the chain of Section~\ref{sec:measured}. Test: the last 30\% of the
unmeasured part in recorded order (1{,}500 and 495 datapoints for the two
arrays); ceiling: an LMMSE trained on its first 70\%.

\begin{table}[!htb]
\centering
\footnotesize
\begin{tabular}{lcc}
\toprule
prior & full grid & per antenna \\
\midrule
generic CDL & 7.12 & 1.81 \\
CDL\,+\,50 from the measured part & 6.11 & 1.66 \\
ray-traced, ITU materials & 12.06 & 2.05 \\
ray-traced, calibrated & 11.63 & 1.54 \\
ray-traced, calibrated, at the exact test positions & 12.05 & 1.68 \\
ray-traced, calibrated\,+\,50 from the measured part & 7.94 & 1.66 \\
CDL\,+\,50 from the unmeasured part (reference) & 2.41 & 0.06 \\
\bottomrule
\end{tabular}
\caption{Calibrated hallway model as a prior where nothing was measured: dB
above the site-trained ceiling at 10\,dB SNR, mean over the two arrays and
over four draws of 50 measurements.}
\label{tab:hallway}
\end{table}

Table~\ref{tab:hallway} gives the result. The main claim, that the calibrated
model beats generic CDL by at least 0.5\,dB on the full grid, failed: it was
4.5\,dB worse. Calibration beat the ITU materials by 0.4\,dB (that claim was
met), and the model lost to CDL with 50 measurements from the measured part by
5.5\,dB. Per antenna, the calibrated model was the best prior without
measurements from the tested part (1.54 against 1.81\,dB), so its power and
delay structure helps a little; its spatial structure does not. At the same
test positions, its normalised spatial signatures matched the measured ones
barely better than an unrelated measured position did (0.45 against 0.39 and
0.74 against 0.76 for the two arrays, post hoc), and knowing the exact test
positions did not help. The model, the surveyed positions and the antenna
phase calibration cannot be separated here (one array's calibration coherence
was 0.55), and the idealised array model is untested. Fifty measurements from
the tested part itself closed almost the whole gap.

\section{Discussion and limitations}

The reversal is consistent with a home-ground effect, which the caveat written
before the first run anticipated: test channels came from the same ray tracer as the
priors. CDL is fitted to measurements; ray tracing at default settings encodes
geometry but, at this site, misses multipath the real street has, and every
ray-traced candidate is biased the same way, so pooling cannot insure against
it. Even closing the gap would not, on its own, be useful: CDL is free and
needs no maps, simulation or tuning, so a ray-traced prior earns its cost only
where it clearly beats CDL. Our simulation study suggested such sites exist
(Florence and San Francisco gained 1--1.5\,dB over CDL in simulation), but the
measured street is not one, and in the follow-on the factory hall's gap was no
larger than the street's; the lab room's larger gap shrinks per antenna once a
short-delay profile is used (Section~\ref{sec:cdlgap}).
Several limits apply.
\begin{itemize}
\item The map-built ray-traced priors were tested on one measured site: one street, one building, three arrays on its
  facade, 1.272\,GHz, mostly line-of-sight; LMMSE estimation only. The
  calibrated hallway model (Section~\ref{sec:hallway}) is one trial on one
  simplified scene (three surfaces, one material each, an idealised array
  model).
\item The measured-channel test was pre-registered but not blind: a pilot had
  already scored CDL on the test data, and a dry run preceded the full run.
\item The simulated placement was not matched to the real users (median 46\,m
  and $\sim$10$^\circ$ below the array by the placement rule, versus
  18--20\,m and 24--26$^\circ$ measured), and one downtilt was used for all
  three arrays.
\item Small asymmetries remain: measured channels sit $\sim$10--20\,ns later
  after alignment than simulated ones (bounded at $\sim$0.5\,dB of the
  CDL-vs-pool gap), and the correlation between the two antenna rows is lower
  in the measured data than in either model, for reasons not established.
\item Most randomly placed users in the rooftop simulations received no paths
  and were dropped, so simulated users skew toward simpler links.
\item The follow-on compares three recordings made with different arrays
  (2$\times$8 on the street and in the lab, 2$\times$4 in the factory), one
  test recording each, ceilings trained differently (separate recordings on
the street, 80\% of one recording indoors) and at different absolute levels,
and indoor site
  measurements from one contiguous stretch of a recording; the lab's
  transmitter height is assumed (positions are two-dimensional).
\item The scattering, distance and tuning probes are exploratory: 7
  cities, scattering levels up to 0.7 (the tuned level sat at the top of that
  range), chosen after seeing earlier results.
\end{itemize}
A code review during the measured-channel work found a cache keyed on object
identity that could hand a scene another scene's ground mesh; it was fixed and
everything rerun, and reported numbers moved by at most 0.02\,dB. A review of
the follow-on found the factory timing offset described in
Section~\ref{sec:cdlgap}; the factory was rerun after the fix.

\section{Conclusion}

In simulation, a covariance prior pooled over many ray-traced cities looked
like a good starting point for an unseen site and a safer one than any single
city. On measured channels from one site it was worse than the 3GPP CDL
models: the map-built scenes produce cleaner channels than the real street,
and neither more cities, diffuse scattering, nor a scattering level tuned from
site measurements closed the gap once users stood at their real distances.
Learned models never beat the classical methods; at best they refined them.
The practical reading is that, without scene content the maps lack, a
ray-traced prior from other cities is not yet a better default than the free
3GPP models. A follow-on on two more measured environments points the same
way: the generic CDL prior was 2.1--4.7\,dB from a site-trained one on the full
grid at all three sites, no larger in a factory hall than on the street and
largest in a small room whose delay spread sits at the low end of the prior's
range; blending 50 site measurements into it closed most of the per-antenna
gap, though not on every array.

\paragraph{Open questions.} (i) \emph{Where does CDL fail?} Three recordings
found no clear case; more environments, recorded with one array type and with
site measurements spread over the area, would be needed to find sites where a
site-specific prior pays off, and only there would simulation be worth its
cost.
(ii) \emph{Missing scene content:} random street clutter (vehicles, poles,
trees) or 3GPP-style random clusters on top of the ray-traced
paths~\cite{tr38901} are the natural next levers. Per-site calibration of a
fully modelled scene is established~\cite{hoydis2023learning}; in a short
final trial (Section~\ref{sec:hallway}) we calibrated that paper's hallway
model on one part of its recording and used it as a prior in another. It was
the best per-antenna prior without local data, but on the full array it was
4.5\,dB worse than generic CDL: the simplified scene does not reproduce the
cross-antenna structure. Transfer of global
settings across sites remains untested. (iii) \emph{More measured sites and
prediction on measured channels}, which need multi-site outdoor arrays and
millisecond-spaced measurements of moving users respectively.

\paragraph{Reproducibility.} Code, the lab notebook with every
pre-registration and result in order, logs and result tables are available
from the author on request. Channel sets are regenerated by the scripts; the measured data is
public under CC BY 4.0~\cite{dichasusdata}.


\begin{thebibliography}{99}\small
\bibitem{tr38901} 3GPP, ``Study on channel model for frequencies from 0.5 to 100 GHz,'' TR 38.901 (v17.0.0, 2022); delay scaling in \S7.7.3.
\bibitem{sionna} J.~Hoydis, S.~Cammerer, F.~Ait~Aoudia, A.~Vem, N.~Binder, G.~Marcus, A.~Keller, ``Sionna: An open-source library for next-generation physical layer research,'' arXiv:2203.11854, 2022.
\bibitem{sionnart} J.~Hoydis, F.~A\"it~Aoudia, S.~Cammerer, M.~Nimier-David, N.~Binder, G.~Marcus, A.~Keller, ``Sionna RT: Differentiable ray tracing for radio propagation modeling,'' arXiv:2303.11103, 2023.
\bibitem{geo2sigmap} Functions Lab, ``geo2sigmap,'' \url{https://github.com/functions-lab/geo2sigmap} (commit 08ca016).
\bibitem{dichasus} F.~Euchner, M.~Gauger, S.~D\"orner, S.~ten~Brink, ``A distributed massive MIMO channel sounder for `big CSI data'-driven machine learning,'' WSA 2021; arXiv:2206.15302.
\bibitem{dichasusdata} University of Stuttgart, ``CSI dataset dichasus-dxxx-complete: Outdoor---three arrays distributed along the facade,'' DaRUS, doi:10.18419/DARUS-3153, CC BY 4.0.
\bibitem{dichasuslab} F.~Euchner, M.~Gauger, ``CSI dataset dichasus-0d5x: Distributed arrays: Indoor LoS, lab room,'' DaRUS, doi:10.18419/DARUS-2218, 2022, CC BY 4.0.
\bibitem{dichasusfactory} F.~Euchner, M.~Gauger, ``CSI dataset dichasus-cf0x: Distributed antenna setup in industrial environment, day 1,'' DaRUS, doi:10.18419/DARUS-2854, 2024, CC BY 4.0.
\bibitem{dichasushallway} F.~Euchner, P.~Stephan, M.~Gauger, ``CSI dataset dichasus-dcxx: Institute hallway with known 3D model, distributed antennas,'' DaRUS, doi:10.18419/DARUS-3831, 2023.
\bibitem{delmoro2026} L.~Del~Moro \emph{et al.}, ``Site geometry and calibration uncertainties in digital twin-enabled channel estimation,'' arXiv:2607.09334, 2026.
\bibitem{alikhani2025} S.~Alikhani \emph{et al.}, ``Digital twin aided channel estimation: Zone-specific subspace prediction and calibration,'' arXiv:2501.02758, 2025.
\bibitem{abouamer2024} M.~S.~Abouamer \emph{et al.}, ``Prediction of wireless channel statistics with ray tracing and uncalibrated digital twin,'' arXiv:2411.13360, 2024.
\bibitem{sifo2026} C.-J.~Zhao \emph{et al.}, ``SiFo: Wireless foundation model for low-overhead site-specific CSI feedback,'' arXiv:2605.16141, 2026.
\bibitem{wifo} B.~Liu \emph{et al.}, ``WiFo: Wireless foundation model for channel prediction,'' arXiv:2412.08908, 2024.
\bibitem{hoydis2023learning} J.~Hoydis, F.~A\"it~Aoudia, S.~Cammerer, F.~Euchner, M.~Nimier-David, S.~ten~Brink, A.~Keller, ``Learning radio environments by differentiable ray tracing,'' arXiv:2311.18558, 2023.
\bibitem{stenhammar2023} O.~Stenhammar, G.~Fodor, C.~Fischione, ``A comparison of neural networks for wireless channel prediction,'' arXiv:2308.14020, 2023.
\bibitem{morais2025} J.~Morais, A.~Malhotra, S.~Hamidi-Rad, A.~Alkhateeb, ``Comparing stochastic and ray-tracing datasets in machine learning for wireless applications,'' arXiv:2512.12449, 2025.
\end{thebibliography}
\end{document}